\documentclass[11pt, A4paper]{article}

\usepackage{lastpage}
\usepackage[colorlinks, linkcolor=blue]{hyperref}
\usepackage{fancyhdr}

\usepackage{color, colortbl}
\usepackage{wrapfig,booktabs}
\usepackage{graphicx}
\usepackage{subcaption}
\usepackage{float}
\usepackage[export]{adjustbox}
\usepackage{wrapfig}
\definecolor{Gray}{gray}{0.9}
\begin{document}
\title{\textbf{Machine Learning based Pallets Detection and Tracking in AGVs}\vspace{-3em}}
\date{04-04-2020 \vspace{-1em}}
\maketitle
\begin{center}

\author{\textbf{Shengchang Zhang}$^{[1]}$, \textbf{Jie Xiang}$^{[2]}$,\textbf{Weijian Han}$^{[2]}$\\
([1]szhang63@ukt.edu,University of Tennessee, Knoxville. [2]\{jxiang19,hanwj\}@stanford.edu, Stanford University)}
\end{center}


\begin{abstract}

The use of automated guided vehicles (AGVs) has played a pivotal role in manufacturing and distribution operations, providing reliable and efficient product handling. In this project, we constructed a deep learning-based pallets detection and tracking architecture for pallets detection and position tracking. By using data preprocessing and augmentation techniques and experiment with hyperparameter tuning, we achieved the result with 25\% reduction of error rate, 28.5\% reduction of false negative rate, and 20\% reduction of training time.
\vspace{-1em}
\end{abstract}

\section{Introduction and Related Work}
Automated Guided Vehicles (AGVs) have been used in distribution, fulfillment, and manufacturing for many years to improve operational efficiency and address shortages in labor. 
In the past several years, AGVs have been widely used in factories and warehouses to locate and track objects. However, the object detection methods currently used in the industry still have room for improvement in terms of efficiency and accuracy.


Ihab S Mohameda et al. \cite{2dlaser, PDT} describe experiments done on the detection and localization of pallets using data collected by a 2D laser rangefinder. Their research provides us with a dataset of 565 2D scans from real-world shop-floor and warehouse environments.

Several state-of-the-art approaches have achieved excellent performance in real applications.
Some key performance metrics, such as object detection and positioning accuracy, for commercial AGVs based on the traditional magnetic tracking approach still do not meet the requirements of many industrial applications today.\\

In this project, we built a machine learning based pallet detection and tracking architecture. We constructed a  Faster Region-based Convolutional Neural Network (Faster R-CNN) model for pallet detection. Data preprocessing and augmentation is applied in model training to improve model accuracy and generalizability. We also constructed a CNN-based classifier and trained it to predict whether the captured images contains any pallets. By optimizing training and tweaking hyperparameters, our model reduced the error rate by 25\%, reduced false negative rate by 28.5\%, and reduced the training time by 20\% compared to the baseline model. Finally, we applied the optimized CNN classier to assist the Faster R-CNN model for pallets tracking.
\section{Dataset and Features} 
After researching the topic extensively and comparing between the quality of different data sources, we chose the 2D Laser Rangefinder dataset contributed by Ihab S. Mohamed \cite{PDT},  which contains a total of 565 images of 2D laser scans. We obtain 2D images by converting the range data from polar to Cartesian coordinates and resizing them to 250 by 250 pixels.
 \vspace{0.5cm}
  \begin{figure}[htbp]  
   \begin{minipage}[t]{0.23\linewidth} 
       \centering 
 \includegraphics[scale=0.28]{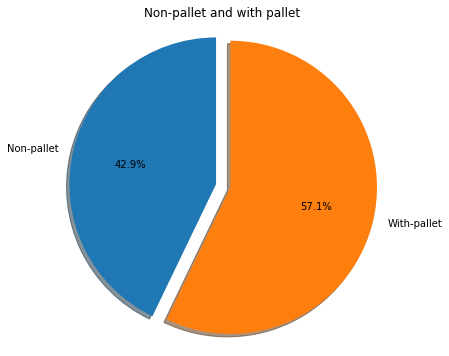}
     \caption{Labeled  }
 \label{}
   \end{minipage}%
   \begin{minipage}[t]{0.25\linewidth} 
     \centering 
   \includegraphics[scale=0.28]{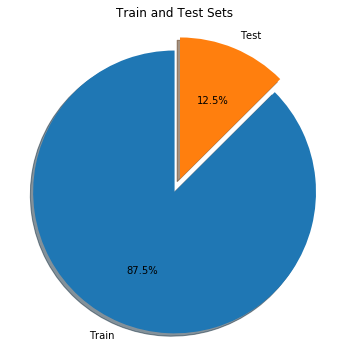}
  \caption{ Distribution }
 \label{datadis}
   \end{minipage} 
   \begin{minipage}[t]{0.2\linewidth} 
     \centering 
   \includegraphics[scale=0.33]{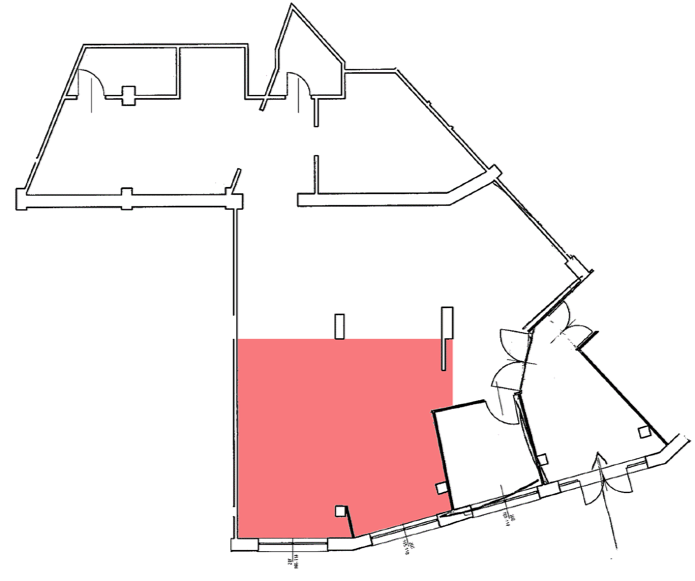}
 \label{datatrac}
   \end{minipage} 
    \begin{minipage}[t]{0.2\linewidth} 
     \centering 
    \includegraphics[scale=0.13]{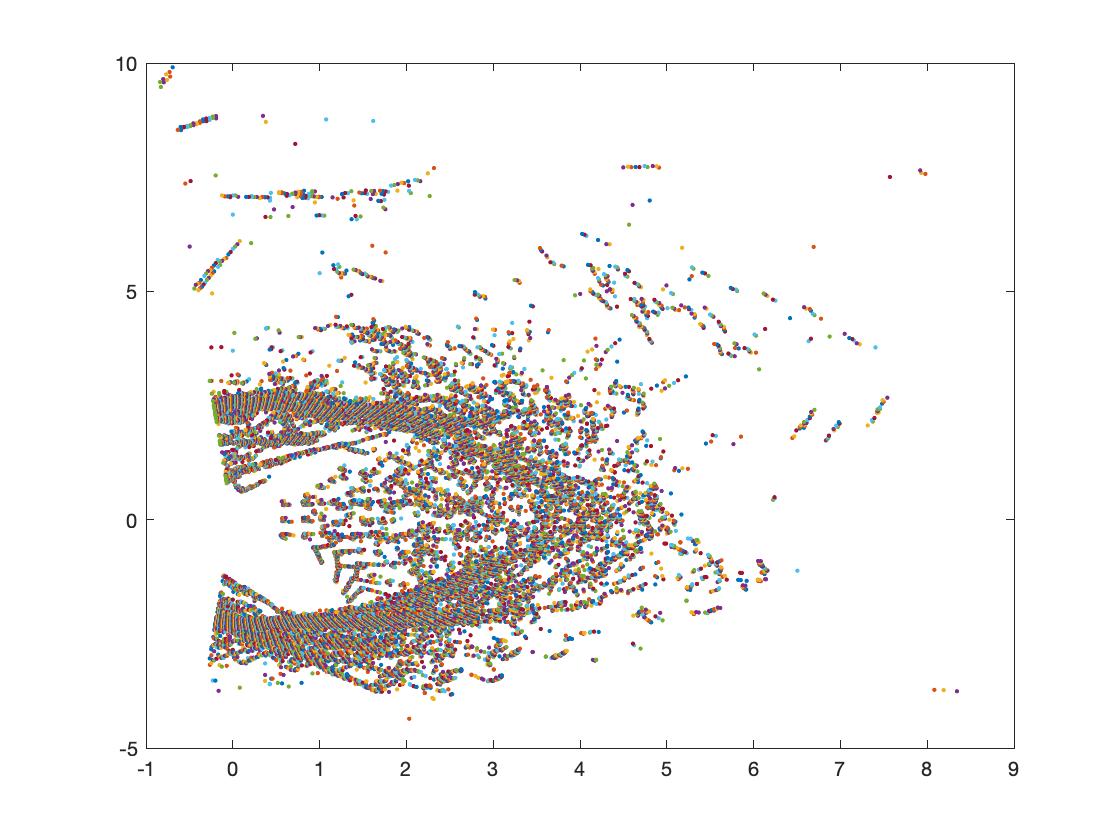}
  \caption{trajectories raw range }
 \label{trackenv}
   \end{minipage} 
 \end{figure}
There are a total of 565 scans, 340 of which contain a pallet, while the remaining 225 do not. In order to train the Faster R-CNN detector, we divide the 340 images containing pallets into two parts: 70\% as the training set and 30\% as the test set.

The data collection process was done in an indoor environment (see Figure\ref{trackenv}).
The 2D laser rangefinder moved along several trajectories inside the red area. 
%
\subsection{Data Processing}
\begin{figure}[htbp]  
   \begin{minipage}[t]{0.5\linewidth} 
       \centering 
   \includegraphics[scale=0.2]{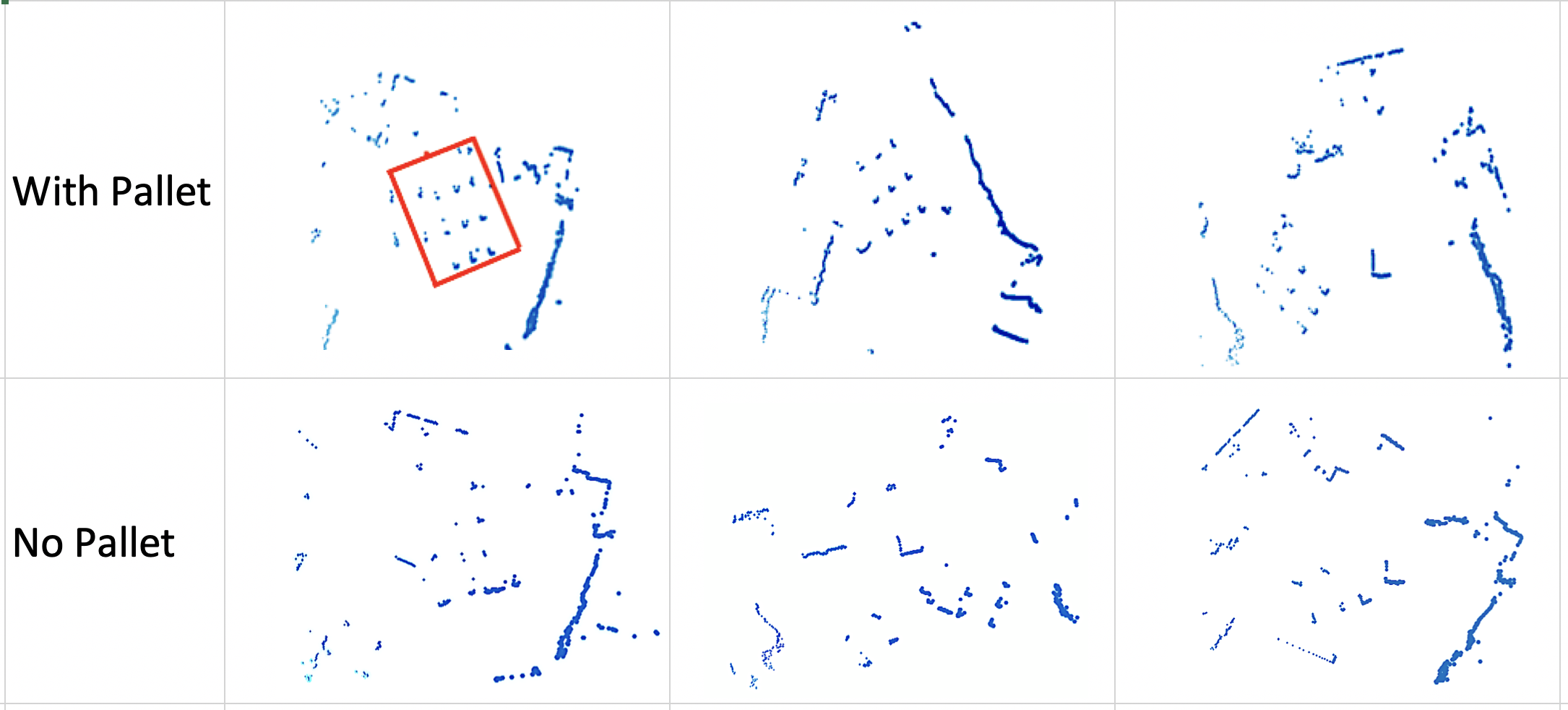}
\caption{Images with pallet or not}
    \label{fig:palletwithno}
    \end{minipage}%
    \begin{minipage}[t]{0.45\linewidth} 
       \centering 
\includegraphics[width=0.9\linewidth]{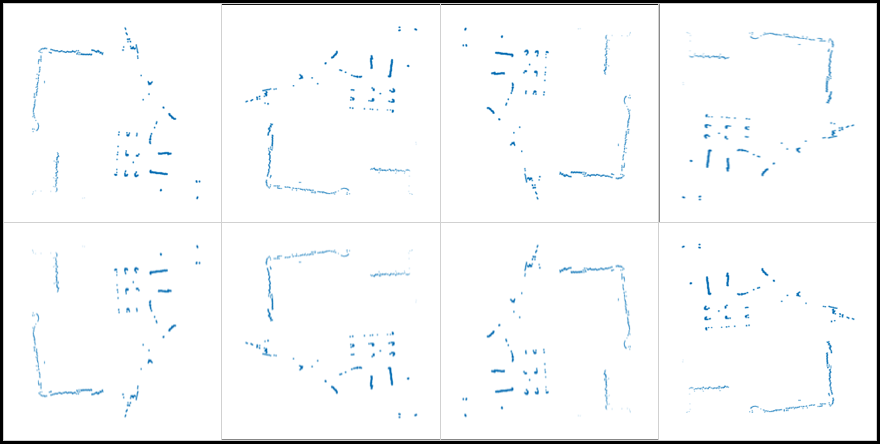}
\captionof{figure}{Data Augmentation}
    \label{fig:augment}
    \end{minipage}%
\end{figure}

The raw range data provided by the laser rangefinder is visualized using the standard ROS package rviz, Fig.\ref{fig:palletwithno} shows examples of the dataset of real-world 2D scans represented in Cartesian coordinates.
The first row shows examples of the dataset where a pallet is present in the environment. The second row shows examples of the dataset when no pallet is present.

\vspace{0.6cm}


\subsection{Data Augmentation}
Because of the shortage of training data, we decided to augment these images to expand our dataset and improve model generalizability. 
For each image, we rotate it by 90, 180, and 270 degrees, and also reflect each image (including the rotated ones) over the x-axis. Using this data augmentation technique, we were able to increase the size of our data set by a factor of 8. and  each image is turned into 8 images as in Fig.\ref{fig:augment}. 
\section{Method}

In order to test the performance of the models used by \cite{2dlaser}, we first construct a Region Proposal Network (RPN) for proposing regions of interest (ROIs) and a Faster Region-based Convolutional Neural Network (Faster R-CNN) for performing object detection on the pallets.
We then construct a CNN-based classifier and train it to predict whether or not images contain pallets, which will be used to assist the Faster R-CNN model in tracking the vehicle.\\
We've improved upon the baseline model and verified  the prediction results, re-architected the network and tuned the hyperparameters.
After verifying the model prediction results, we went ahead and improved on the methodology by optimizing the code infrastructure, model architecture, and training hyperparameters for both the Faster R-CNN and CNN models. We also explored data preprocessing techniques, using different data transformations and augmentations to help refine training.

Then we compared the performance of these models by looking at the prediction accuracy, precision, and recall.\\


\textbf{Pallet Detection Model}

\begin{wrapfigure}{r}{9.5cm}

  \includegraphics[scale=0.9] {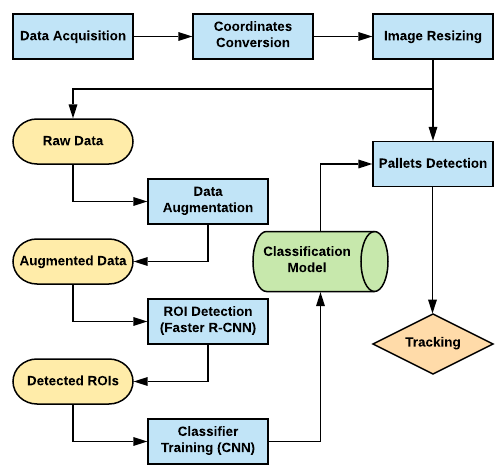}
   \caption{{Pallet Detection System} }\label{fig:picture}
\end{wrapfigure}

The pallet detection process is made-up of two steps: a state-of-the-art Faster R-CNN detector which uses its region proposal network to propose the regions of interest in each image, and a CNN-based classifier taking as input the previous step and determines which of them could be a possible pallet candidate (see \ref{fig:picture}). We then take the input images, preprocess them to be in Cartesian coordinates and the same image size, and then feed them into the Faster R-CNN pallet detection model.

The Faster R-CNN detector is composed of several layers:
the input layer, intermediate hidden layers, and the output layer.

The input layer consists of the input image corresponding to the 2D scan, down-scaled to a $32 \times 32$ pixel grey-scale or RGB images to improve general performance.

For the intermediate hidden layers, there are two convolutional layers, interleaved by two ReLU layers, and followed
by a final max-pooling layer, which produces output images of size $30 \times 30$ pixels.
Each convolutional layer applies up to 25 filters, with a size of 3 and a stride and a padding of 1, whereas the max-pooling layer employs pooling regions of size 3 and a stride of 1.

The final stage is composed of one fully connected layer using the ReLU activation function, and another fully connected layer using the softmax activation function. The first fully connected layer outputs the top 64 most significant features in the image, which are then used by the second fully connected layer to determine whether a ROI proposed by the RPN belongs to one of the object classes or to the background, using sequential classification. The overall output is a list of candidate ROIs.


\section{Experiments, Results, and Discussion}
\subsection{Training }
We experimented with adding and removing layers for the neural networks and adjusting the filters in each layer, and we also tweaked other hyperparameters such as the learning rate, number of folds for k-fold cross validation, and the number of training epochs.
SGD and k-fold cross-validation (with k = 2, 3, 5, 8, 10) are used to train the CNN-based classifier with an initial learning rate 
$\alpha = 0.001, 0.005, 0.01, 0.03, 0.05, 0.1, $
and mini-batch size set to 50, leading to the following data.
Considering both performance and computation time, we selected learning rate = 0.1, max epochs = 10, folds of cross validation = 5,  number of filters = 15, and convolutional layers = 1 in the final model. See in Table \ref{tab-parameter}.\\




\newlength{\oldintextsep}
\setlength{\oldintextsep}{\intextsep}

\setlength\intextsep{-1pt}
\begin{wraptable}{r}{8cm}
\caption{Hyperparameters Tuning}
\label{tab-parameter}
\centering
 \scalebox{0.75}{\begin{tabular}{||c | c | c | c||} 
 \hline
\cellcolor{Gray}
Learning Rate &\cellcolor{Gray} Accuracy &\cellcolor{Gray} Precision &\cellcolor{Gray} Recall \\ [0.5ex] 
 \hline\hline
 0.1 & 0.981 & 0.990 & 0.974 \\ 
 \hline
 0.05 & 0.980 & 0.990 & 0.972 \\
 \hline
 0.03 & 0.979 & 0.984 & 0.976 \\
 \hline
 0.01 & 0.963 & 0.977 & 0.952 \\
 \hline
 0.005 & 0.953 & 0.961 & 0.948 \\
 \hline
 0.001 & 0.919 & 0.908 & 0.942 \\ [1ex] 
 \hline
  Max Epochs & Accuracy & Precision & Recall \\ [0.5ex] 
 \hline
 3 & 0.969 & 0.982 & 0.960 \\ 
 \hline
 5 & 0.980 & 0.986 & 0.976 \\
 \hline
 10 & 0.986 & 0.994 & 0.980 \\
 \hline
 15 & 0.982 & 0.988 & 0.978 \\ [1ex] 
 \hline
 Folds & Accuracy & Precision & Recall \\ [0.5ex] 
 \hline\hline
 2 & 0.966 & 0.989 & 0.946 \\ 
 \hline
 3 & 0.981 & 0.984 & 0.980 \\ 
 \hline
 5 & 0.988 & 0.990 & 0.988 \\
 \hline
 8 & 0.989 & 0.990 & 0.990 \\
 \hline
 10 & 0.977 & 0.963 & 0.994 \\ [1ex] 
 \hline
 Filters & Accuracy & Precision & Recall \\ [0.5ex] 
 \hline\hline
 5 & 0.938 & 0.942 & 0.940 \\ 
 \hline\
 10 & 0.939 & 0.957 & 0.926 \\ 
 \hline
 15 & 0.948 & 0.965 & 0.936 \\
 \hline
 20 & 0.943 & 0.959 & 0.932 \\
 \hline
 25 & 0.940 & 0.964 & 0.920 \\ [1ex] 
 \hline
 Layers & Accuracy & Precision & Recall \\ [0.5ex] \hline\hline
 1 & 0.940 & 0.966 & 0.918 \\ 
 \hline
 2 & 0.893 & 0.868 & 0.938 \\ 
 \hline
 3 & 0.508 & 0.526 & 0.666 \\
 \hline
\end{tabular}}
\end{wraptable}
\textbf{Learning Rate}
 We tried using multiple different learning rates for training, and found that having learning rates smaller than 0.1 actually do not improve our model's performance. The source model used a constant learning rate of 0.03, and we were able to improve the model's performance as well as reduce training time by increasing the learning rate to 0.1.\\
 \textbf{Max Epochs}
From the training metrics measured on the average of all the folds in cross validation, we conclude that the model likely converges with around 5 epochs. The source model trains with 10 epochs, which has very minimal improvements in all three metrics (accuracy, precision, and recall). Training for 15 epochs clearly overfits because generalization is worse than with 10 epochs on all three metrics. \\
\textbf{Numbers of Folds}
 We found that all three metrics are consistently high for 3, 5, and 8 folds, but the variation is much larger with 2 and 10 folds. 
 we conclude that keeping 5 folds of cross validation is a good approach, because the model performance improvement over 3 folds is significant, but the additional performance improvement by increasing it to 8 folds is negligible. However, when we are tuning hyperparameters with many options, we choose to use 3-fold cross validation to save on computation time.\\
\textbf{Number of Filters}
Using the same hyperparameter tuning methodology as before, we see that using 15 filters for the convolutional layer performs the best among all the options we tried. The source model used 25 filters, but our selection improves upon it in all three metrics, with an especially large improvement in recall.\\
\textbf{Number of Layers}
We discovered that the source model does in fact have the most reasonable number of layers, because adding one additional layer does not add much to model performance, and adding two additional layers significantly overfits the training data, leading to very poor generalization.
 \subsection{Model Evaluation}
\begin{figure}[htbp]  
   \begin{minipage}[t]{0.43\linewidth} 
       \centering 
   \includegraphics[scale=0.16]{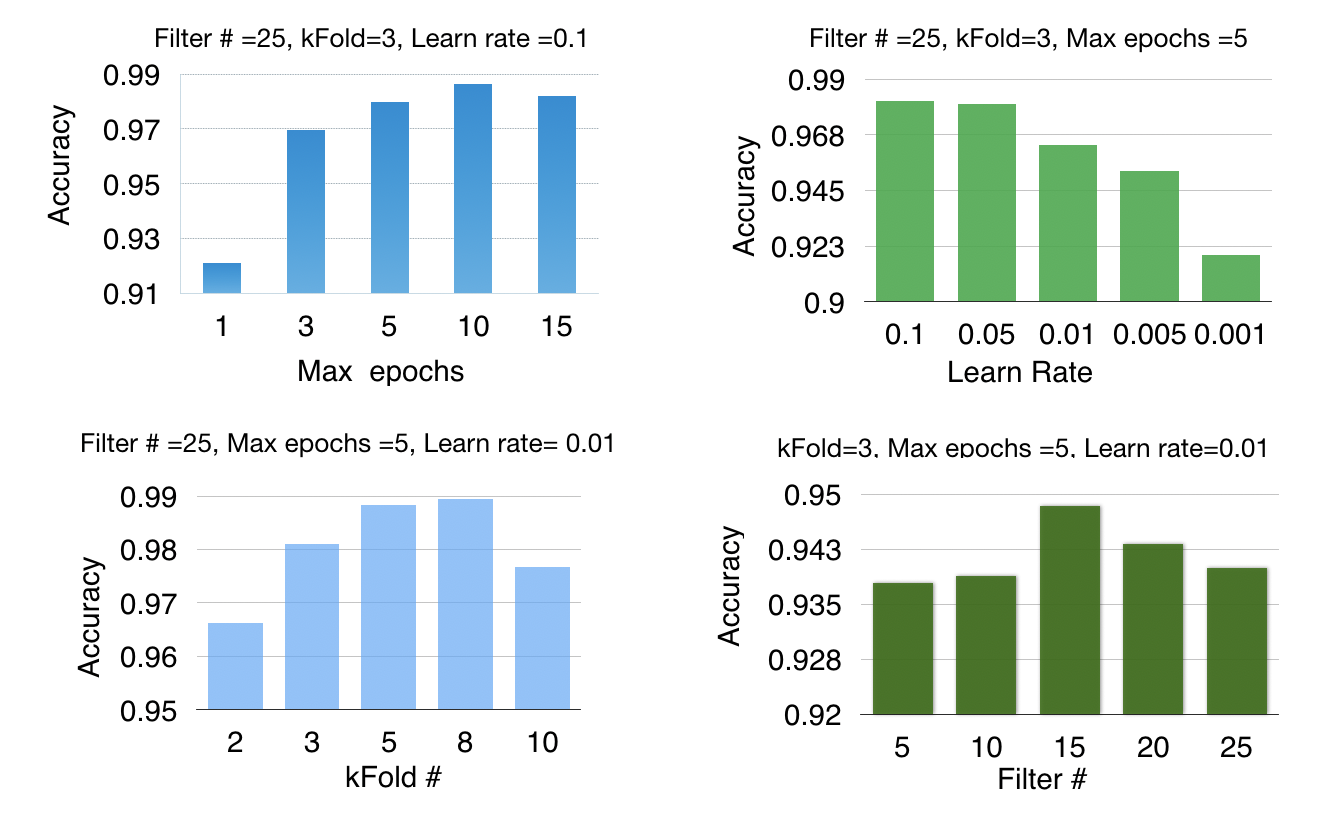}
\caption{Parameters tuning}
    \label{fig:parameter}
    \end{minipage}%
    \begin{minipage}[t]{0.55\linewidth} 
       \centering 
\includegraphics[scale=0.45]{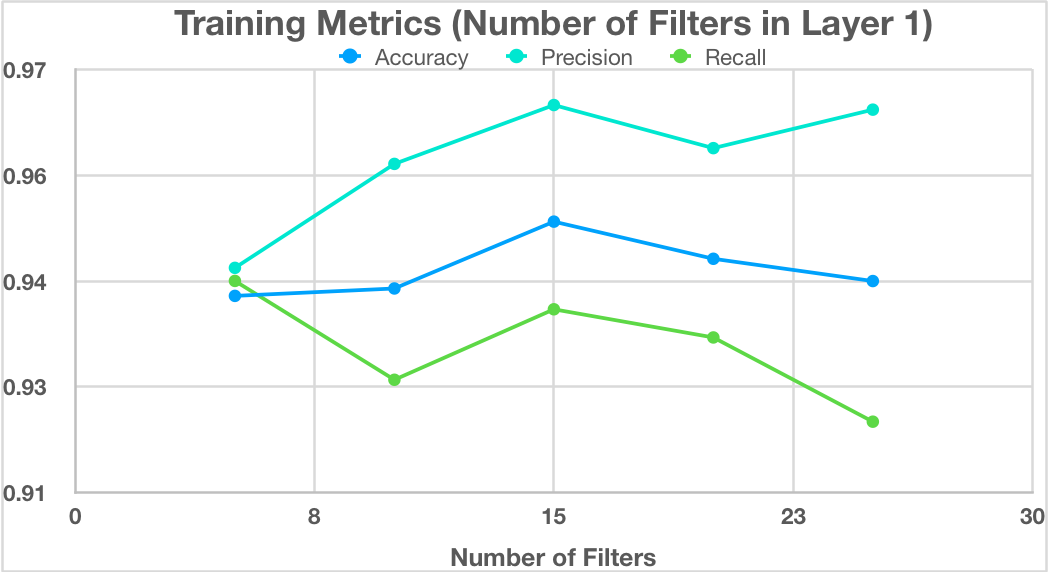}
\captionof{figure}{Accuracy, Precision, and Recall}
    \label{fig:APR}
    \end{minipage}%
\end{figure}

Figure \ref{fig:parameter} shows  examples of how we selected the best number of parameters to use in the first hidden layer of the CNN model. For example, we selected 15 filters because it offers a good balance of model complexity and performance, and using it gives us a higher accuracy, precision, and recall compared against the baseline model (see Figure \ref{fig:APR}) .
\vspace{0.6cm}
\vspace{1cm}

We compared the performance of our optimized model against the original model by looking at the prediction accuracy, precision, and recall on the test set.
Training our optimized model on the training set, and testing on the test set yielded the following results: accuracy = 0.994, precision = 0.998, and recall = 0.990. and also took 20\% less time to train.This represents an error rate reduction of 25\%, which is very significant when it comes to guiding autonomous vehicles, where the last 1\% of edge cases are often the hardest to overcome.

\section{Conclusion and Future Work}

Utilizing machine learning techniques, we can assist AGVs in detecting pallets and tracking their locations more accurately and with less latency, thus improving their operational safety and efficiency. Our model architecture improvements, data preprocessing and augmentation, and hyperparameter tuning helped us optimize the Faster R-CNN model and CNN-based classifier, reducing the error rate by 25\%, the false negative rate by 28.5\%, and training time by 20\%.
\\

In order to implement this deep learning based pallet detection and tracking systems for AGVs in the warehouse, the following scenario need to be considered and implemented, which will be our future work:\\
(1) Pallet orientation estimation
and pallet position estimation accuracy improvement
\\
(2) Pallet type classification
and multiple pallets detection
\\
(3) Improve the AGV’s efficiency by using reinforcement learning so that the AGV can learn the shortest route towards a targeted pallet.

\newpage
\section{Contribution }
All three of us contributed significantly to the methodology, research, model reproduction, and analysis of results of this project.
On top of that, each individual also had the following contributions:
{Shengchang Zhang} selected the research topics and guided the direction of research and the final report.
{Weijian Han} constructed the model and tuned hyperparameters,  analysis training results. Weijian also performed data preprocessing and data augmentations.
{Jie Xiang} trained the CNN models, tuned hyperparameters, and improved the infrastructure of the training code.
\\
And we would also like to thank the CS 229 course instructors for teaching ML course, sections and homework, especially TA Ethan Steinberg, and Jingbo Yang for giving us guidance in the project.


\begin{thebibliography}{0}

\bibitem{2dlaser}Ihab S Mohameda,Alessio Capitanellib, Fulvio Mastrogiovannib, Stefano Rovettab,
Renato Zaccariab. A 2D laser rangefinder scans dataset of standard EUR pallets.arXiv:1805.08564v2 [cs.RO] 13 Mar 2019
\bibitem{PDT} Ihab S Mohamed, Alessio Capitanelli, Fulvio
Mastrogiovanni, Stefano Rovetta, Renato Zaccaria. 
Detection, localisation and tracking of pallets using machine learning techniques and 2D range data. arXiv:1803.11254v3 [cs.RO] 28 Apr 2019
\bibitem{pdtdata}Data: \url{https://github.com/EmaroLab/PDT/tree/master/Palle-Detection/AllData}
\bibitem{Vservoing}Vignesh S, Rajesh Kanna S K, Lingaraj N.Intelligent Automated Guided Vehicle using Visual Servoing.American Journal of Engineering Research (AJER) e-ISSN: 2320-0847 p-ISSN : 2320-0936 Volume-6, Issue-11, pp-16-20
\bibitem{ssd}Wei Liu, Dragomir Anguelov, Dumitru Erhan, Christian Szegedy, Scott Reed, Cheng-Yang Fu, Alexander C. Berg. SSD: Single Shot MultiBox Detector. arXiv:1512.02325v5 29 Dec 2016
\bibitem{yolo}Joseph Redmon, Santosh Divvala, Ross Girshick, Ali Farhadi. You Only Look Once: Unified, Real-Time Object Detection. arXiv:1506.02640v5 9 May 2016


\end{thebibliography}
\end{document}